\documentclass[lettersize,journal]{IEEEtran}
\usepackage{amsmath,amsfonts}
\usepackage{amssymb}
\usepackage{algorithm}
\usepackage{array}
\usepackage[caption=false,font=normalsize,labelfont=sf,textfont=sf]{subfig}
\usepackage{textcomp}
\usepackage{stfloats}
\usepackage{url}
\usepackage{verbatim}
\usepackage{graphicx}
\usepackage{cite}
\usepackage{subfig}
\usepackage{gensymb}
\usepackage{algpseudocode}

\begin{document}

\title{CTP: A hybrid CNN-Transformer-PINN model for ocean front forecasting}

\author{
	\IEEEauthorblockN{Yishuo Wang$^{a,b}$, Feng Zhou$^{b,a,*}$, Muping Zhou$^{b}$, Qicheng Meng$^{b}$, Zhijun Hu$^{c,b}$, Yi Wang$^{c,b}$}
	
	\IEEEauthorblockA{$^a$ School of Oceanography, Shanghai Jiao Tong University, China}
	
	\IEEEauthorblockA{$^b$ State Key Laboratory of Satellite Ocean Environment Dynamics, Second Institute of Oceanography, MNR, China}
	
	\IEEEauthorblockA{$^c$ College of Marine Science and Technology, China University of Geosciences, China}
	
	\IEEEauthorblockA{\{Yishuo Wang\}wys1998@sjtu.edu.cn, \{Feng Zhou\}zhoufeng@sio.org.cn, \{Muping Zhou\}zhoumuping@sio.org.cn, \{Qicheng Meng\}q.meng@sio.org.cn, \{Zhijun Hu\}huzhijun@cug.edu.cn, \{Yi Wang\}1202311356@cug.edu.cn}
}

\maketitle

\begin{abstract}
This paper proposes CTP, a novel deep learning framework that integrates convolutional neural network(CNN), Transformer architectures, and physics-informed neural network(PINN) for ocean front prediction. Ocean fronts, as dynamic interfaces between distinct water masses, play critical roles in marine biogeochemical and physical processes. Existing methods such as LSTM, ConvLSTM, and AttentionConv often struggle to maintain spatial continuity and physical consistency over multi-step forecasts. CTP addresses these challenges by combining localized spatial encoding, long-range temporal attention, and physical constraint enforcement. Experimental results across south China sea(SCS) and Kuroshio(KUR) regions from 1993 to 2020 demonstrate that CTP achieves state-of-the-art(SOTA) performance in both single-step and multi-step predictions, significantly outperforming baseline models in accuracy, $F_1$ score, and temporal stability.
\end{abstract}


\section{Introduction}
\IEEEPARstart{O}{cean} fronts, characterized by sharp gradients in physical and biogeochemical properties such as temperature, salinity, and nutrient concentrations, are critical yet dynamic features of the global ocean\cite{Yang-2024}. These transitional zones, formed by the convergence of distinct water masses, play a pivotal role in regulating energy transfer, material cycling, and biological processes across marine ecosystems\cite{Woodson-2015}. The study of fronts is essential for advancing understanding of ocean dynamics, as they act as hotspots for vertical mixing, influence large-scale circulation patterns, and modulate air-sea interactions that impact regional and global climate systems\cite{Champon-2020}. Ecologically, fronts enhance primary productivity by facilitating nutrient upwelling, thereby supporting trophic cascades and aggregating diverse marine species, from phytoplankton to apex predators. Furthermore, fronts are increasingly recognized for their socioeconomic relevance, as they shape fisheries productivity, guide migratory routes of commercially important species, and affect the dispersion of pollutants in a changing ocean. Automatic detection, tracking, and prediction of ocean fronts are therefore essential.

Front detection is handled by many previous researchers \cite{Yang-2016, Canny-1986, Prants-2014, Xing-2023, Cayula-1992}. This study is based on the detection results of  microcanonical multiscale formalism(MMF) model and tracking results of metric space analysis\cite{Tamim-2015,Wang-2025}.  In addition, researchers have typically focused on predicting other oceanographic variables such as sea surface temperature(SST) and chlorophyll with relatively limited efforts devoted to ocean front prediction.

Due to the lack of image understanding, traditional machine learning methods are not suitable for this study\cite{Hu-2020, Guo-2021}. Therefore, deep learning methods are used and compared in this paper. In the past, long short-term memory(LSTM) is usually adopted for prediction task not only in natural language processing but also in oceanography\cite{Russwurm-2020, Sadeghifar-2017, Xiao-2019, Su-2021}. Spatial features are also added by CNN to form convolutional LSTM(ConvLSTM) algorithm\cite{Zhang-2020,Zhou-2024,Petrou-2019,Zhang-2020}. However, Transformer have become dominant in prediction task nowadays with the advantage of attention mechanism,  which has boosted the development of large language models(LLMs) and ocean elements prediction\cite{Attention-2017,Gao-2023,Dai-2024,Yue-2024}.Yang combined CNN and attention mechanism to form AttentionConv framework for ocean front prediction\cite{Yuting-2024}. PINN add constraint equations for these networks to regularize and disclose the dynamics of ocean phenomena\cite{Meng-2023,Brunton-2020,Cui-2022}. This work proposes CTP framework to incorporate physical information to improve performance. Through this, spatial, temporal and physical information are well blended to form a more robust and precise model.

\subsection*{Key advantages over prior work}
1. Physics-informed regularization: PINN constraints introduce domain knowledge through governing equations such as Navier-Stokes(N-S), enhancing physical plausibility and reducing overfitting.

2. Superior predictive performance: Extensive evaluations show that CTP consistently outperforms LSTM, ConvLSTM, CNN-LSTM-PINN(CLP), and AttentionConv in terms of accuracy, $F_1$ score, and robustness across diverse regions and time scales.

3. Balanced computational cost: Despite its hybrid structure, CTP achieves competitive training efficiency, making it suitable for large-scale, long-term forecasting tasks in operational oceanography.

\section{Proposed CTP Architecture}
\subsection{Data}
The SST is obtained from the NOAA Coral Reef Watch daily global $5$km($0.05\degree$ exactly) satellite coral bleaching heat stress monitoring product(v3.1). This fine-grained resolution is essential for understanding ocean fronts and it can be downloaded from \url{https://coralreefwatch.noaa.gov/product/5km/}. 

The eastward and northward velocity($u$ and $v$ respectively) is from the global ocean physics reanalysis product released by Copernicus marine service. The spatial resolution is $9$km($1/12\degree$ exactly) and can be downloaded from \url{https://data.marine.copernicus.eu/product/GLOBAL_MULTIYEAR_PHY_001_030/services}.

SCS and KUR are selected as regions studied in this paper. The spatial range is $100-125\degree$E, $0-25\degree$N and $120-145\degree$E, $20-45\degree$N respectively, with the temporal range 1993-2020. The resolution is unified to $9$km, so the spatial dimension is $300\times300$ and $10227$ daily records are available. In this paper, the task is to predict the frontal zone, $u$, $v$ using frontal zone, $u$, $v$ in the preceding one week. Therefore, the input dimension is $7\times3\times300\times300$, the target is $1\times3\times300\times300$ and the total number of samples is $10220$.

\subsection{CNN}
When it comes to prediction task, each sample has a spatial size of $3\times300\times300=270000$. If such a large amount of data is rearranged into a one-dimensional vector, it will lead to two issues: (1) loss of spatial structure, and (2) increased computational cost.

A CNN encoder is placed before the Transformer module in this work. It is composed of $2$ layers and the decoder counterpart includes $2$ deconvolution layers to restore data after Transformer. ReLU is selected as the activation function, group normalization(GN) and transposed convolution(ConvTran) are applied. The data size after CNN is $32\times75\times75=180000$, only $2/3$ of raw size. The number of CNN layers($2$) is determined in \ref{sec:cnn_layers}.

\subsection{Transformer}
Transformer framework in this paper consists of encoders only to reduce calculation complexity. The number of encoder layers is $2$ and the dimension of model is $d_{model}=512$. Multi-head attention mechanism is adopted and the number of headers is $h=8$. Here, the dimension of key, query and value is the same($d_k=d_q=d_v=d_{model}/h=64$) and the dimension of feed-forward network is $1024$.

\begin{equation}\label{eq:Attention}
	Z_i = softmax(\frac{Q_iK_i^T}{\sqrt{d_k}})V_i
\end{equation}

The $7\times3\times300\times300$ initial tensor is reduced to $7\times32\times75\times75$ after the aforementioned CNN decoder , flattened to $7\times180000$ and projected to $7\times d_{model}$ by a linear layer. Then it is multiplied by $W_i^Q\in \mathbb{R}^{d_{model}\times d_q}$, $W_i^K\in \mathbb{R}^{d_{model}\times d_k}$ and $W_i^V\in \mathbb{R}^{d_{model}\times d_v}$ respectively to generate $Q_i$, $K_i$ and $V_i$ for these heads, where $i$ is the index of the specific head. Attention mechaism is imposed using \eqref{eq:Attention} to obtain $Z_i \in \mathbb{R}^{7 \times d_v}$. These $Z_i$ are concatenated to $Z \in \mathbb{R}^{7 \times d_{model}}$. Finally, $Z$ goes through the feed-forward network which is composed of $2$ fully-connected layers and a ReLU function. The network extend $Z$ to $7 \times 1024$, activate it by ReLU and contract it to $7 \times d_{model}$. 

The mean along the temporal dimension of the Transformer output is used as the final prediction embedding, with size $d_{model}$. It is projected to $180000$ and rearranged to $32\times75\times75$. The CNN decoder mentioned above is applied to restore data to size $3 \times 300 \times 300$. In addition, the first component in dimension $2$ should be the probability of frontal zone, so a Sigmoid function is needed to make the value fall in the $(0, 1)$ interval.

\subsection{PINN}
\begin{equation}\label{eq:NS}
	\frac{\partial \vec{v}}{\partial t} + (\vec{v} \cdot \nabla)\vec{v} = \nu \Delta \vec{v} - \frac{1}{\rho}\nabla \vec{p} + \vec{f}
\end{equation}

The N-S equation for momentum conservation is illustrated by \eqref{eq:NS}. In it, $\vec{v}$ represents $u$ or $v$, $\nu$ is kinematic viscosity and its dimension is $m^2/s$, $\rho$ is the water density, $\vec{p}$ is pressure and $\vec{f}$ is the external force per unit mass.

For the items in \eqref{eq:NS}, $\frac{\partial \vec{v}}{\partial t}$ is the rate of change of velocity, $(\vec{v} \cdot \nabla)\vec{v}$ is the convection item and $\nu \Delta \vec{v}$ represents the diffusion. These three terms are incorporated into the loss function other than cross entropy(CE) loss for frontal zone classification and mean squared error(MSE) for velocity prediction. 

\begin{equation}\label{eq:CE}
	CE = -\frac{\sum_{i=1}^{H}\sum_{j=1}^{W}[p_{ij}ln(\hat{p}_{ij})+(1-p_{ij})ln(1-\hat{p}_{ij})]}{HW}
\end{equation}

\begin{equation}\label{eq:MSE}
	MSE = \frac{\sum_{i=1}^{H}\sum_{j=1}^{W}(y_{ij}-\hat{y}_{ij})^2}{HW}
\end{equation}

In \eqref{eq:CE}, $H$ represents the height of studied region and $W$ is the width. $p_{ij}$ is the label of frontal zone($1$ is frontal zone point and $0$ is non frontal zone point) at position $i, j$, $\hat{p}_{ij}$ is the predicted probability at that point. Whereas in \eqref{eq:MSE}, $y_{ij}$ is the target value and $\hat{y}_{ij}$ is the predicted value. $y_{ij}$ can represent $u$, $v$, or the derived physical terms from Eq. \eqref{eq:NS}, such as $\frac{\partial \vec{v}}{\partial t}$, $(\vec{v} \cdot \nabla)\vec{v}$ and $\nu \Delta \vec{v}$.

\begin{equation}\label{eq:loss}
	\begin{aligned}
	loss_1 = &CE(frontal \ zone) + MSE(u) + MSE(v) + \\
	&MSE(\frac{\partial u}{\partial t}) + MSE(\frac{\partial v}{\partial t}) + MSE((\vec{v} \cdot \nabla)u) + \\
	&MSE((\vec{v} \cdot \nabla)v) + MSE(\nu \Delta u) + MSE(\nu \Delta v)
	\end{aligned}
\end{equation}

To sum up, the loss function is composed of $5$ items: CE of frontal zone and MSE of velocity, rate of change, convection and diffusion. CE is suitable for classification task for it is based on Bernoulli distribution and MSE for regression based on Gaussian. In addition, MSE is derivable compared to other choice like mean absolute error(MAE), so the rate of convergent is faster with MSE. It should be clarified that the rate of change and convection is approximated by forward finite difference in time and space respectively. As to diffusion, the second derivative is approximated using central finite difference. The time step($1$ day), spatial resolution($9$km) and $\nu$ ($10^{-6}m^2/s$) should be considered to ensure all physical terms have consistent units($m/s^2$), enabling balanced gradient contributions and stable training. Further demonstration will be shown in \ref{sec:loss}.

\begin{figure*}
	\centering
	\includegraphics[width=1.0\textwidth]{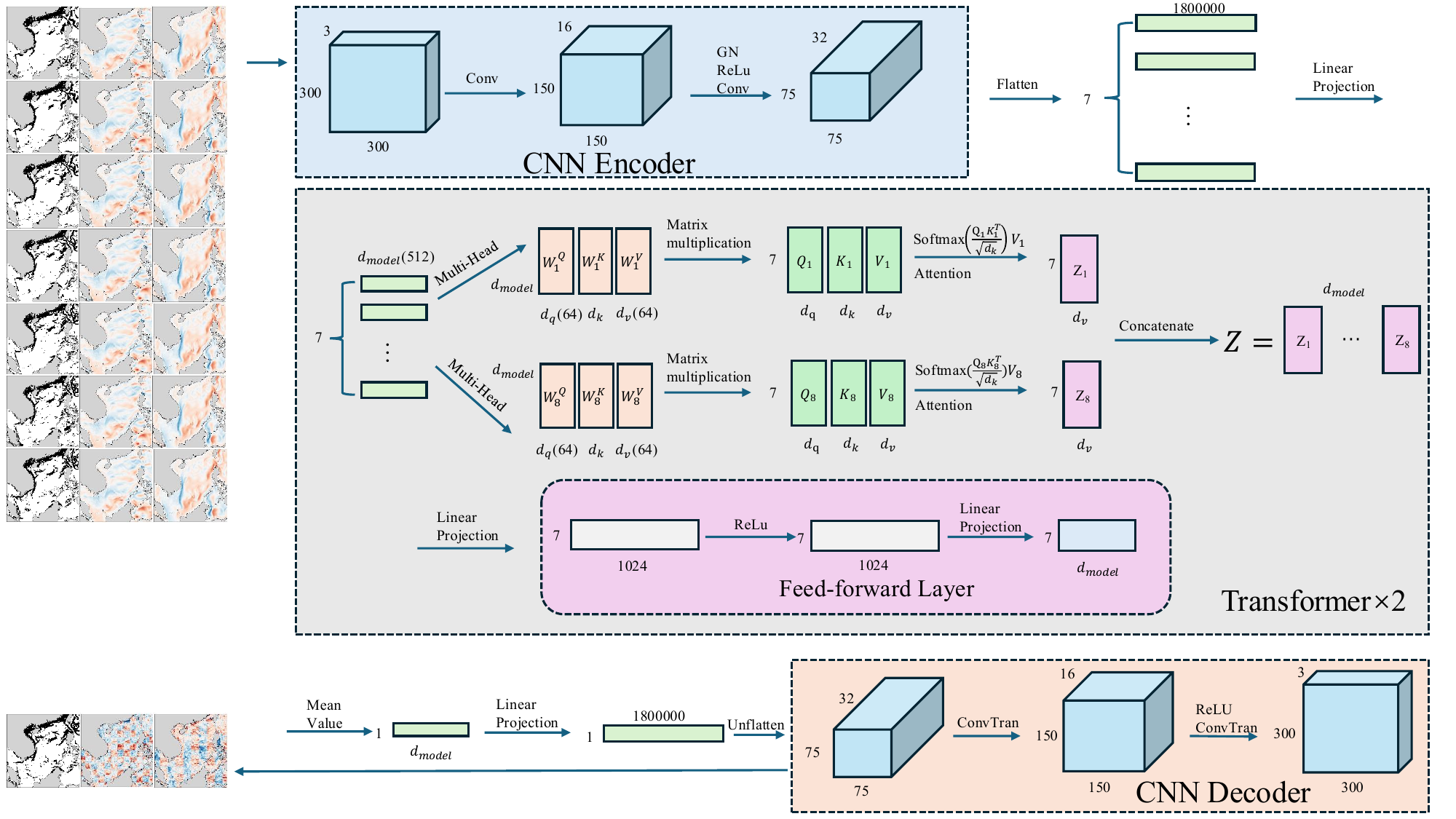}
	\caption{Overall architecture of the proposed CTP framework, including CNN encoding, temporal attention and physical loss integration} 
	\label{fig:flowchart} 
\end{figure*}

The architecture of the proposed CTP framework is illustrated in figure \ref{fig:flowchart}.

\section{Experiments}
\subsection{Experimental Setup}
These hyperparameters: batch size is set at $32$, learning rate is $10^{-4}$ and epoch is $50$ for further comparison. Adam optimizer is used. The ratio of training set and test set is $8:2$, i.e., $8176:2044$ samples. 

In \ref{sec:hyperpara} and \ref{sec:ablation}, only $1/4$ data($2555$ samples) are used to determine parameters and architecture for quick calculation and SCS is selected.

In \ref{sec:compare}, the CLP architecture is also included in the comparison to thoroughly evaluate the effectiveness of different model combinations.. The learning rate of LSTM-based models(LSTM, ConvLSTM and CLP) is $10^{-3}$ for LSTM's recurrent feature. Hidden size is $512$ and number of layers is $2$. In addition, the CNN architecture of ConvLSTM and CLP is the same as CTP.
AttentionConv is the same as Yang's setting\cite{Yuting-2024}.

The device for computing is one NVIDIA Tesla P100 GPU with $16$ GB memory. 

\subsection{Hyperparameter analysis}
\label{sec:hyperpara}

\subsubsection{CNN layers}
\label{sec:cnn_layers}

The number of CNN decoder and encoder layers are changed from $1$ to $4$ to explore its impact on the whole network, with the number of Transformer layers set at $2$ and $loss_1$ used. The tensor size after each layer is $3\times300\times300$, $16\times150\times150$, $32\times75\times75$, $64\times38\times38$ and $128\times19\times19$.

\begin{figure}
	\centering
	\includegraphics[width=0.5\textwidth]{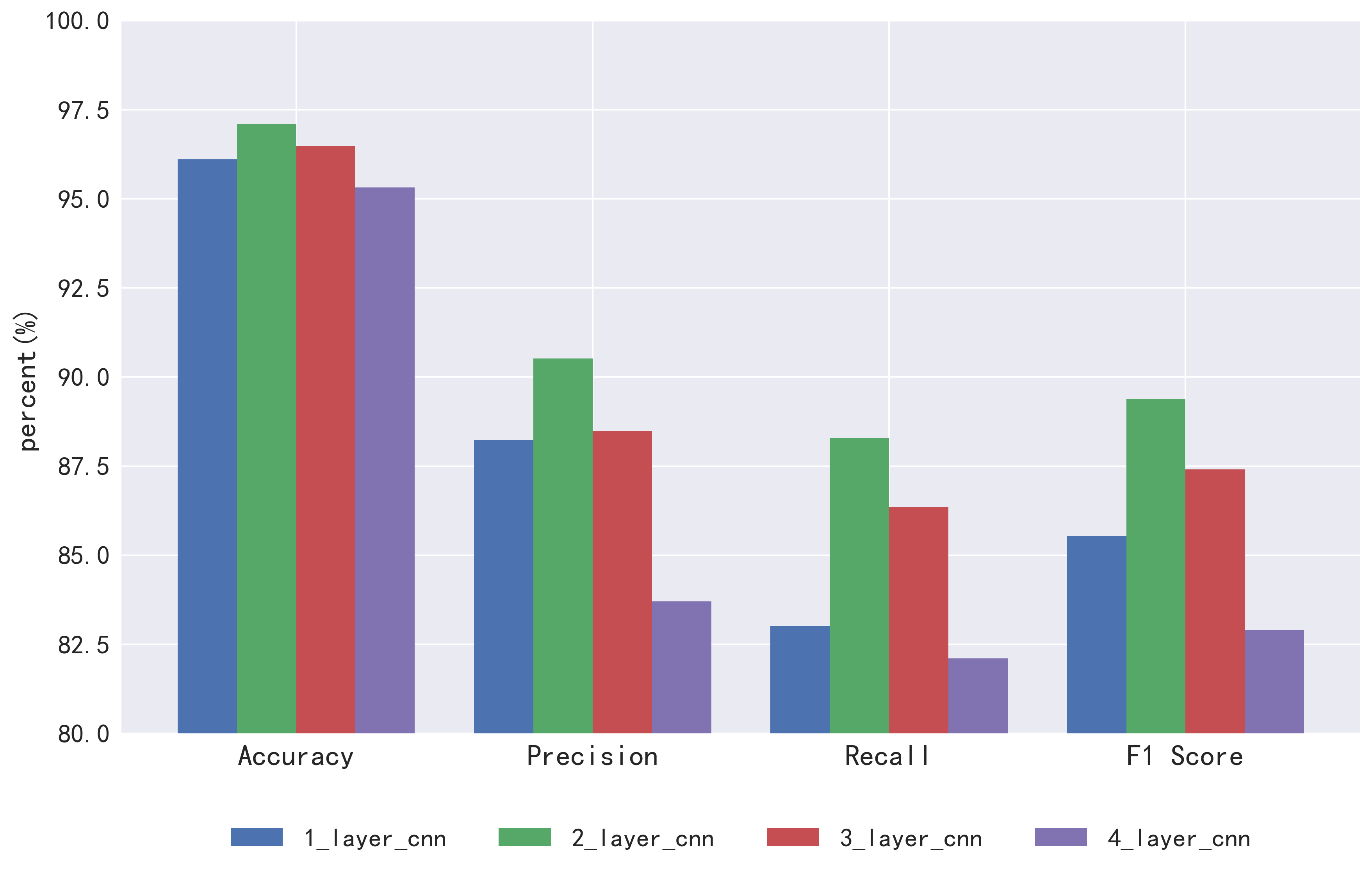}
	\caption{Performance of different number of CNN layers} 
	\label{fig:cnn_layer} 
\end{figure}

The performance comparison across varying layer depth reveals a non-linear relationship between depth and classification metrics, as shown in figure \ref{fig:cnn_layer}. The $2$-layer CNN achieved the best overall performance, with the highest accuracy ($97.10\%$), precision ($90.52\%$), recall ($88.29\%$), and $F_1$ score ($89.39\%$). While $1$-layer and $3$-layer models showed comparable accuracy ($96.11\%$ and $96.48\%$, respectively), their precision and recall were lower, especially for the 1-layer model, indicating limited feature extraction capability. Notably, the $4$-layer CNN performed worst across all metrics, suggesting that excessive depth leads to overfitting or vanishing gradients, resulting in degraded generalization.

\subsubsection{Transformer layers}
With $2$-layer CNN and $loss_1$, the number of Transformer layers is also studied by changing from $1$ to $4$.

\begin{figure}
	\centering
	\includegraphics[width=0.5\textwidth]{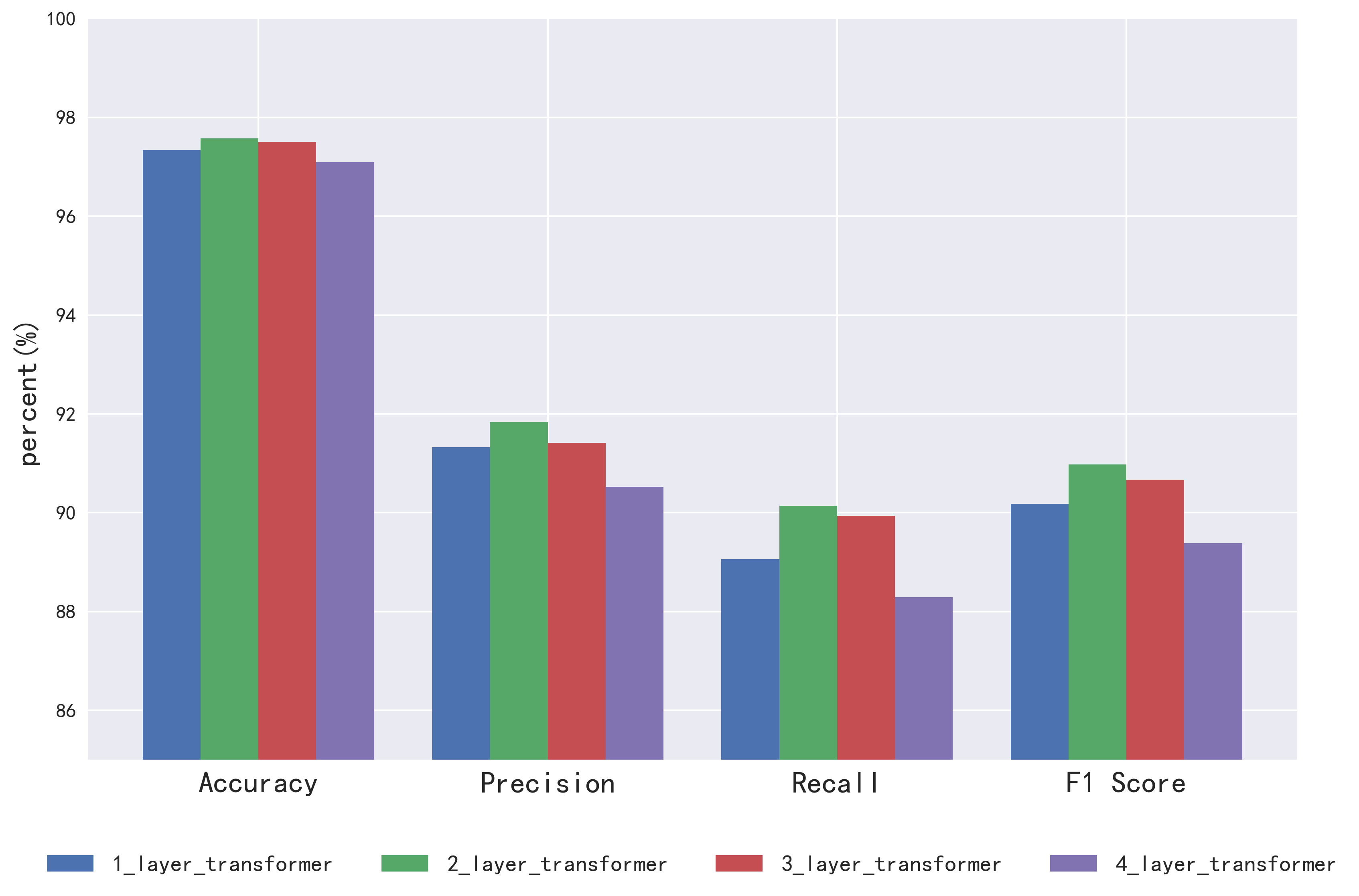}
	\caption{Performance of different number of Transformer layers} 
	\label{fig:trans_layer} 
\end{figure}

The evaluation of Transformer depth shows that performance generally peaks at $2$ layers in figure \ref{fig:trans_layer}. The $2$-layer Transformer achieved the highest scores across all metrics, including accuracy ($97.58\%$), precision ($91.84\%$), recall ($90.14\%$), and $F_1$ score ($90.98\%$). While $1$-layer and $3$-layer models performed similarly, their slight drop in recall and $F_1$ suggests less effective context modeling. The $4$-layer Transformer consistently yields the lowest performance, indicating that deeper architectures may introduce overfitting or optimization difficulties.

\subsubsection{Loss function}
\label{sec:loss}
\begin{equation}\label{eq:MAE}
	MAE = \frac{\sum_{i=1}^{H}\sum_{j=1}^{W}|y_{ij}-\hat{y}_{ij}|}{HW}
\end{equation}

\begin{equation}\label{eq:losses}
	\begin{aligned}
		loss_2 = &MSE(frontal \ zone) + MSE(u) + MSE(v) + \\
		&MSE(\frac{\partial u}{\partial t}) + MSE(\frac{\partial v}{\partial t}) + MSE((\vec{v} \cdot \nabla)u) + \\
		&MSE((\vec{v} \cdot \nabla)v) + MSE(\nu \Delta u) + MSE(\nu \Delta v)\\
		loss_3 = &CE(frontal \ zone) + MAE(u) + MAE(v) + \\
		&MAE(\frac{\partial u}{\partial t}) + MAE(\frac{\partial v}{\partial t}) + MAE((\vec{v} \cdot \nabla)u) + \\
		&MAE((\vec{v} \cdot \nabla)v) + MAE(\nu \Delta u) + MAE(\nu \Delta v)\\	
		loss_4 = &MAE(frontal \ zone) + MAE(u) + MAE(v) + \\
		&MAE(\frac{\partial u}{\partial t}) + MAE(\frac{\partial v}{\partial t}) + MAE((\vec{v} \cdot \nabla)u) + \\
		&MAE((\vec{v} \cdot \nabla)v) + MAE(\nu \Delta u) + MAE(\nu \Delta v)\\				
	\end{aligned}
\end{equation}

The loss function is illustrated by \eqref{eq:loss}, where CE is adopted for frontal zone classification and MSE for the other items. Another commonly used loss fuction is mean absolute error(MAE), displayed by \eqref{eq:MAE}. A natural question arises which loss fucntion assembly is better for this work. $4$ combinations are put forward: CE + MSE, MSE, CE + MAE, MAE, as shown in \eqref{eq:losses}.

\begin{figure}
	\centering
	\includegraphics[width=0.5\textwidth]{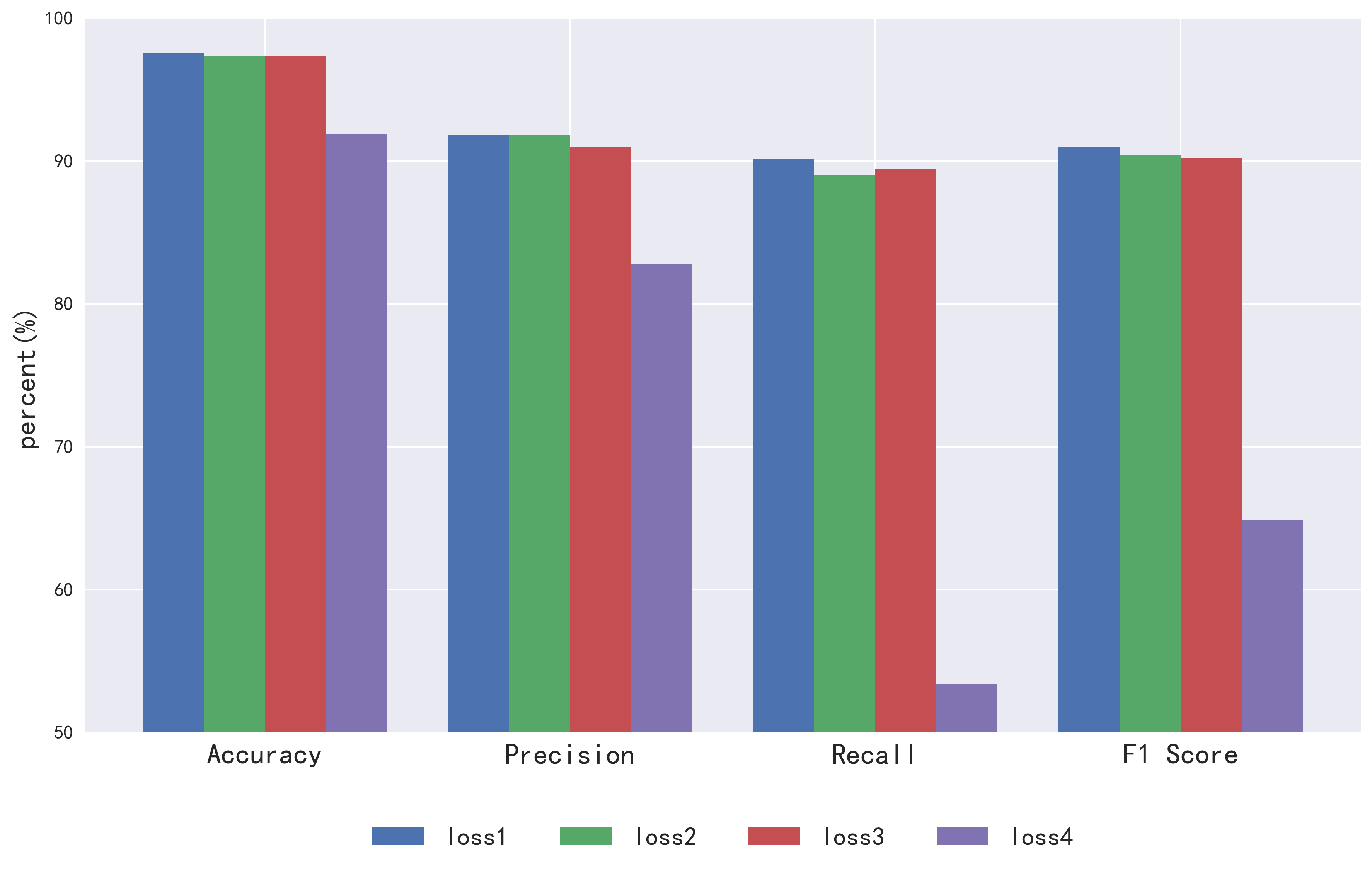}
	\caption{Performance of different loss functions} 
	\label{fig:loss} 
\end{figure}

In figure \ref{fig:loss}, the result($2$-layer CNN and Transformer) shows that $loss_1$, which combines CE for classification and MSE for velocity and physical terms, achieves the best overall performance across all metrics. $loss_2$, which replaces CE with MSE for the frontal zone term, performs comparably but slightly lower, suggesting CE better guides classification. $loss_3$, which uses MAE instead of MSE but retains CE, results in moderate performance, indicating MAE may reduce sensitivity to outliers but lacks the precision of MSE. $loss_4$, which relies entirely on MAE, shows a significant drop in recall and $F_1$ score—especially recall ($53.36\%$)—highlighting that MAE alone is insufficient for capturing the full complexity of both classification and physical constraints in this task.

\subsection{Ablation studies}
\label{sec:ablation}

To investigate the influence of PINN on the proposed network, a model without PINN is trained. The $3$-channel input and $3$-channel output are altered to frontal zone-input and frontal zone-output. The architecture is the same as the former, only the first CNN encoder is $1\times300\times300$ to $16\times150\times150$ and the last decoder layer is $16\times150\times150$ to $1\times300\times300$. The loss function is thus reduced to CE(frontal zone) only.

The CNN module can also be removed to study its function. In this situation, the input $3\times300\times300$ is directedly flattened to $270000$ rather than $180000$. If both CNN and PINN are removed, the input $1\times300\times300$ is flattened to $90000$ before Transformer.

\begin{figure}
	\centering
	\includegraphics[width=0.5\textwidth]{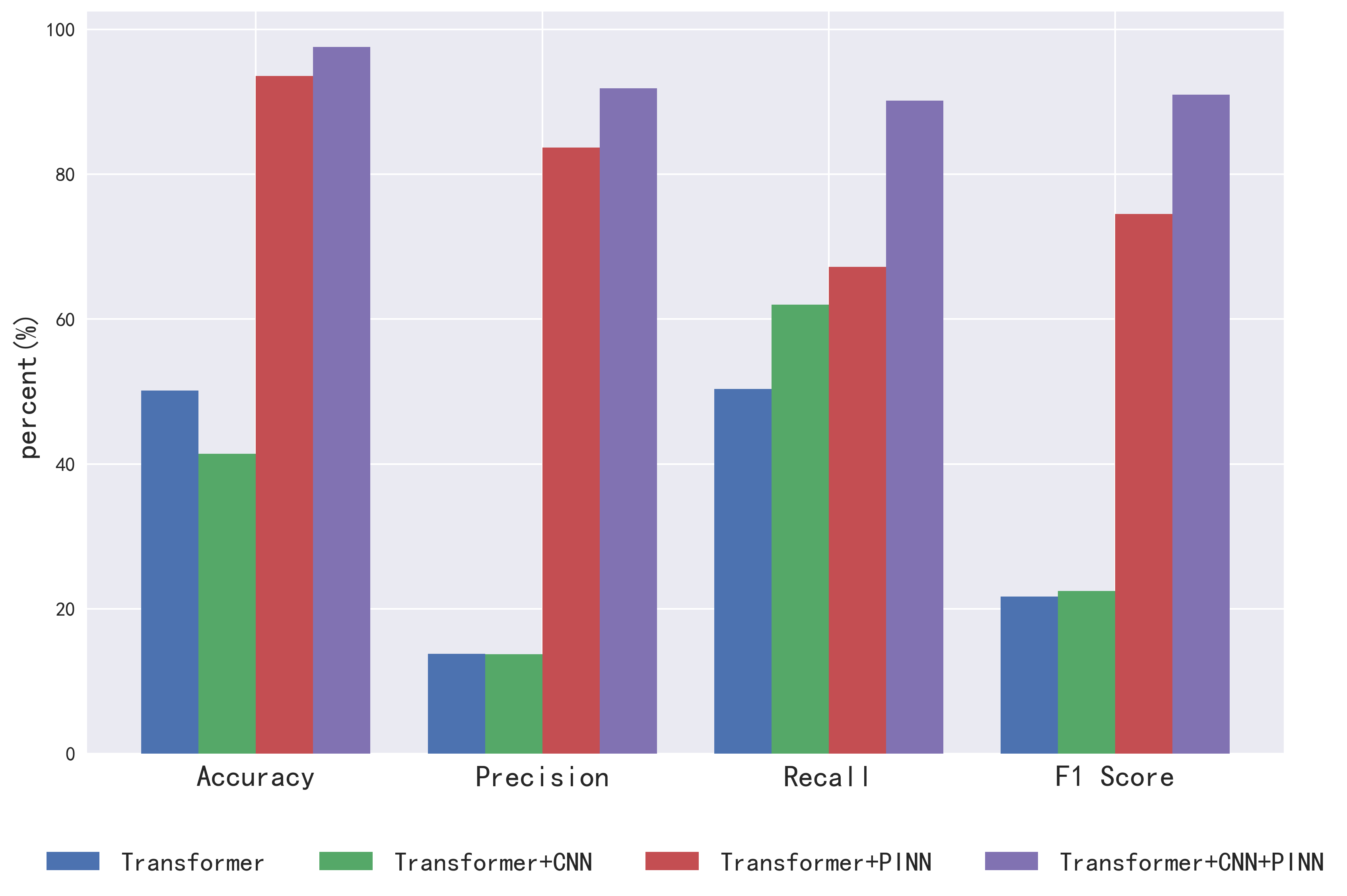}
	\caption{Results of ablation studies} 
	\label{fig:ablation} 
\end{figure}

As displayed in figure \ref{fig:ablation}, the Transformer baseline shows poor overall performance (accuracy: $50.12\%$, $F_1$: $21.67\%$), with particularly low precision ($13.80\%$) suggesting significant false positives. This indicates inadequate feature discrimination. Adding CNN layers slightly reduces accuracy ($41.42\%$) but improves recall ($61.97\%$), implying CNN enhances pattern detection at the cost of increased misclassifications.

The Transformer+PINN configuration achieves a dramatic improvement (accuracy: $93.57\%$, $F_1$: $74.54\%$), where PINN's physics-informed constraints likely stabilize training and enforce domain-specific priors, boosting both precision ($83.66\%$) and recall ($67.21\%$).

The CTP model delivers optimal results (accuracy: $97.58\%$, $F_1$: $90.98\%$). The synergy arises from: CNN extracting localized spatial features; Transformer capturing sequential dependencies; PINN regularizing outputs via physical knowledge.

PINN contributes most significantly to performance gains by mitigating overfitting and aligning predictions with physical plausibility. In addition, CNN alone degrades precision due to overemphasis on local patterns without global context. The full hybrid architecture balances local/global feature learning and domain constraints, achieving harmonized precision ($91.84\%$) and recall ($90.14\%$).

\subsection{Quantative comparisons}
\label{sec:compare}
\subsubsection{Case analysis}
\begin{figure*}
	\centering
	\includegraphics[width=1.0\textwidth]{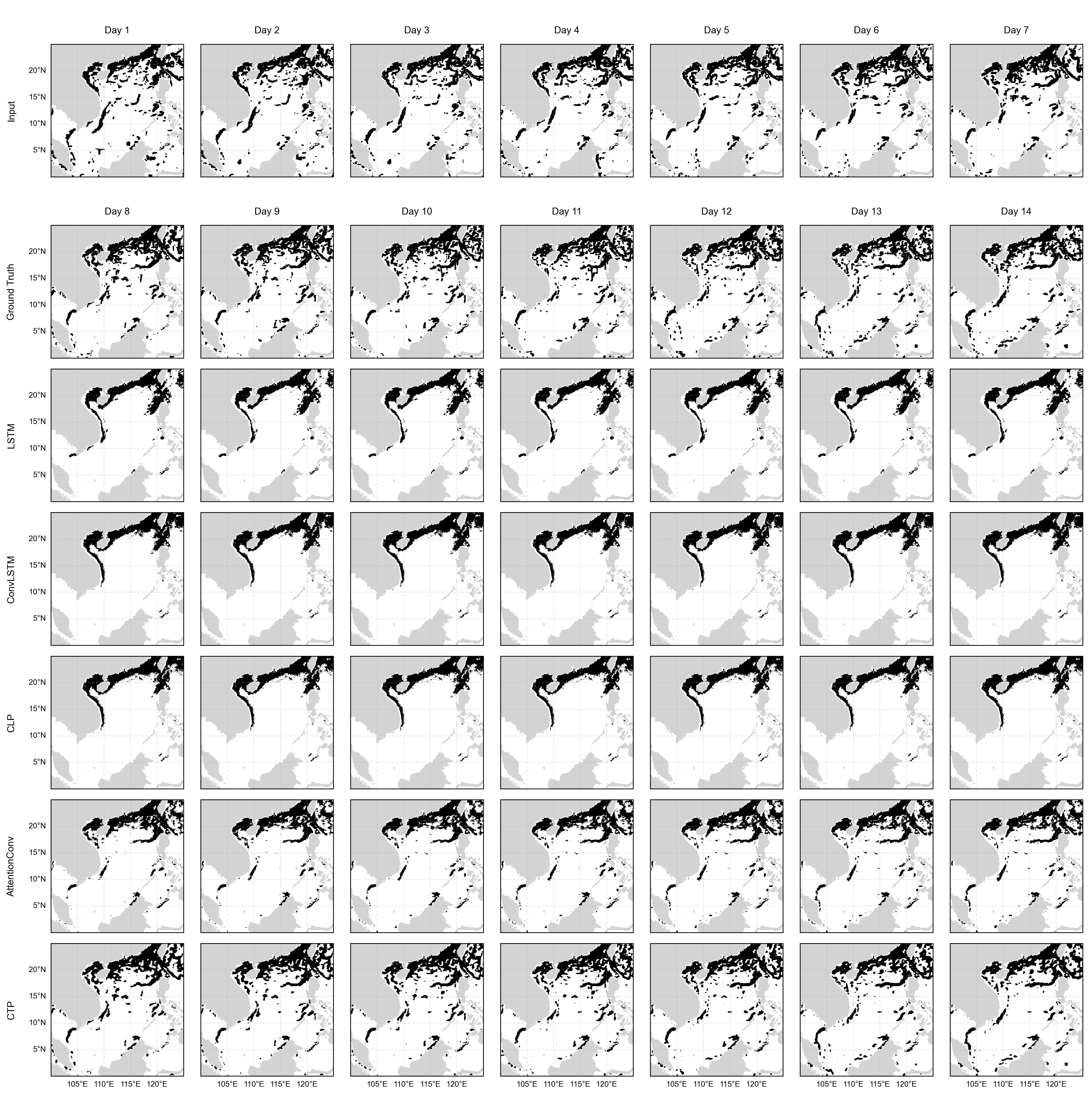}
	\caption{Performance of different methods on SCS in 2020.01.01-2020.01.14} 
	\label{fig:geo_SCS} 
\end{figure*}

\begin{figure*}
	\centering
	\includegraphics[width=1.0\textwidth]{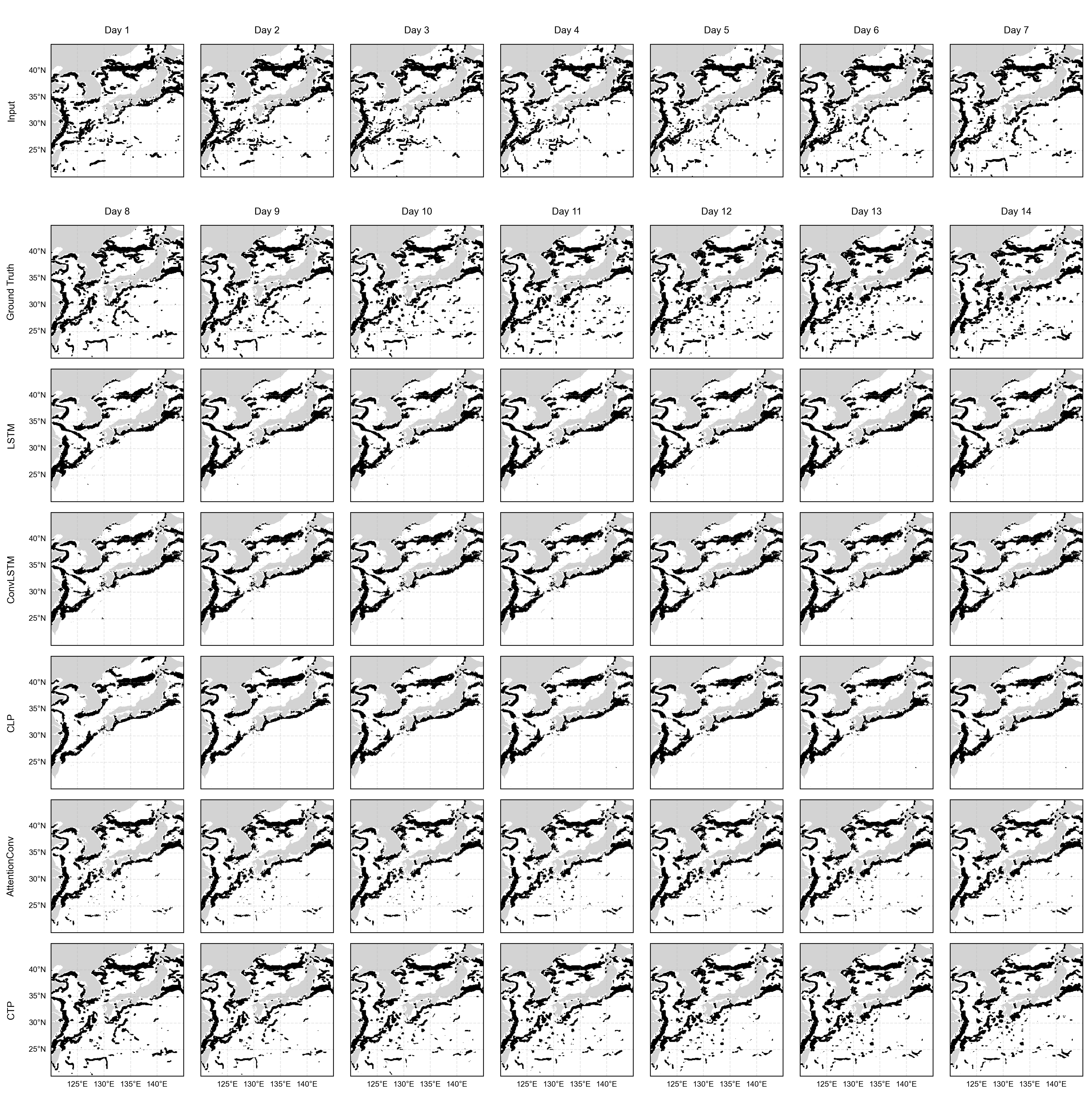}
	\caption{Performance of different methods on KUR in 2020.01.01-2020.01.14} 
	\label{fig:geo_KUR} 
\end{figure*}

Figure \ref{fig:geo_SCS} and \ref{fig:geo_KUR} are the $7$-step prediction cases on SCS and KUR test set. Across both regions, CTP exhibits the most consistent and physically plausible evolution of the frontal structures over the $14$-day sequence. The transitions between days are smooth and well-aligned with the ground truth ($2$nd row), preserving both spatial continuity and fine-scale features, particularly after Day $10$.

The LSTM model captures coarse-scale frontal features initially, but its predictive capability deteriorates in later days. A significant issue is that from around Day $9$ onwards, the output becomes static. This plateau effect indicates LSTM’s difficulty in maintaining dynamic progression across multiple steps—a known limitation when error accumulation and internal state degradation occur in sequence-to-sequence models; ConvLSTM's Incorporating spatial convolutions improves structural recognition compared to LSTM, especially in earlier forecast days. However, the model still fails to sustain the fine-scale features after Day $11$, and frontal transitions become blurred, particularly in SCS; CLP benefits from physical constraints via PINN and produces more realistic outputs. Yet, it still struggles with sharpness and stability, particularly in the KUR, where the frontal zone appears fragmented by the end of the sequence.

AttentionConv improves by applying attention mechanisms to convolutional features. As a result, AttentionConv preserves spatial coherence and enhances detail retention better than ConvLSTM and CLP. However, while it shows clear improvements, it occasionally exhibits patchy noise or over-smoothing in areas with rapid frontal shifts, suggesting limitations in how it handles long-range dependencies and complex physical transitions without explicit physical constraints; CTP combines the strengths of CNN for spatial encoding, Transformers for long-range temporal attention, and PINN for physics-informed regularization. It delivers the most temporally consistent and spatially accurate results across all days. Notably, CTP maintains mesoscale structures and shows minimal degradation even on Day $14$. The transitions in both SCS and KUR are smooth, realistic, and physically plausible.

\subsubsection{$1$-step prediction}
\begin{figure}
	\centering
	\includegraphics[width=0.5\textwidth]{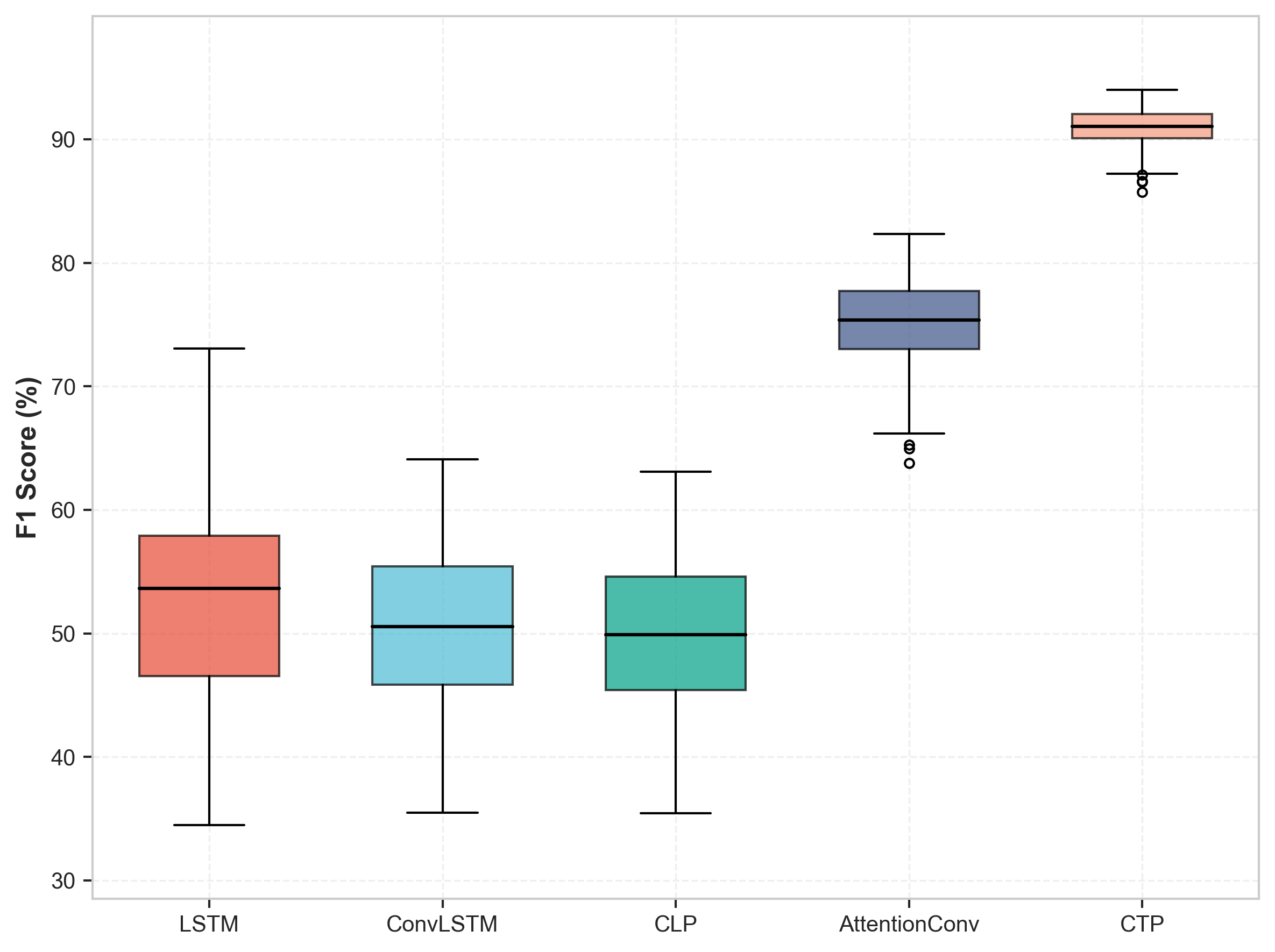}
	\caption{$F_1$ score of different methods on SCS test set($1$-step)} 
	\label{fig:box_SCS} 
\end{figure}

\begin{figure}
	\centering
	\includegraphics[width=0.5\textwidth]{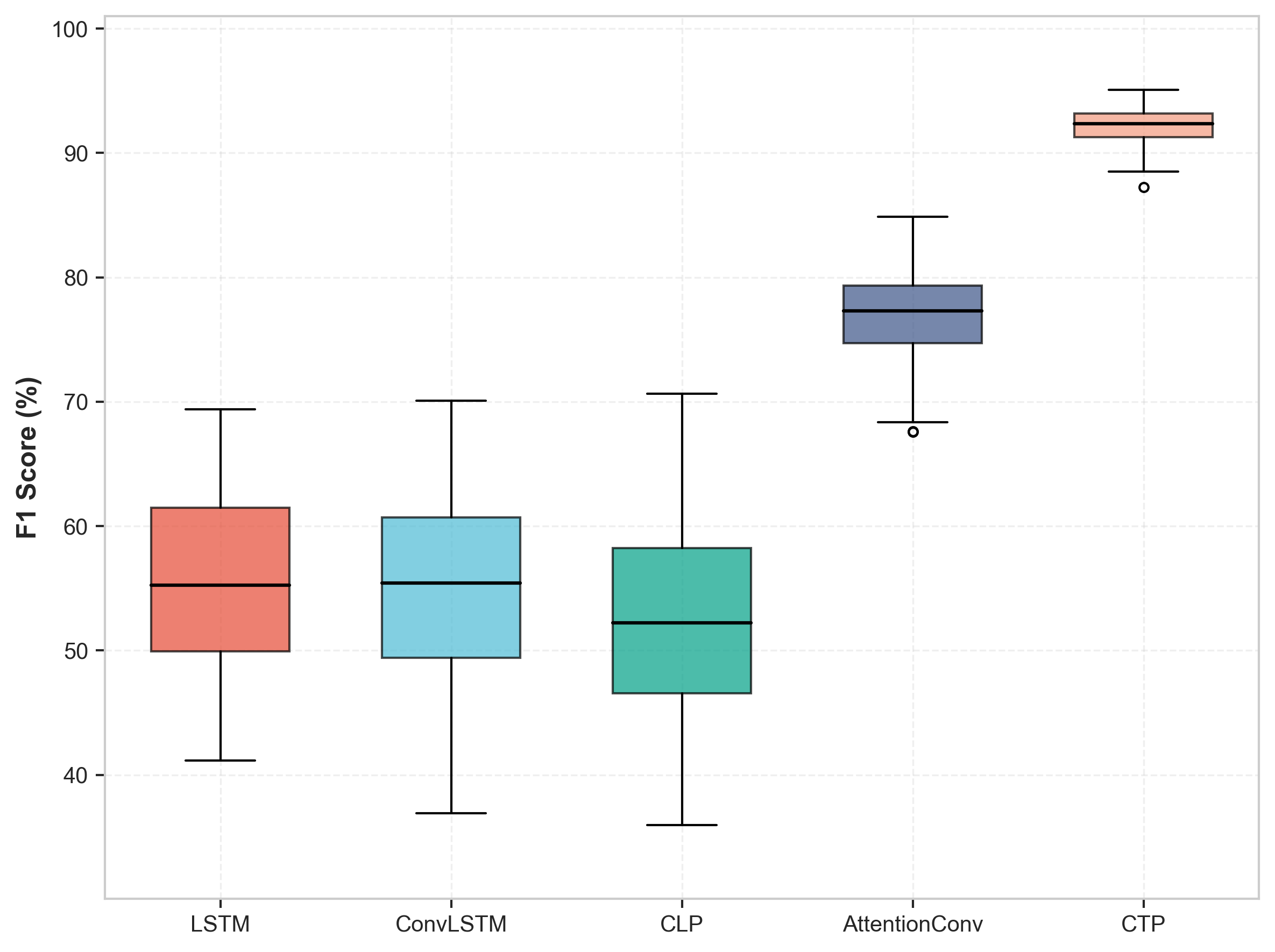}
	\caption{$F_1$ score of different methods on KUR test set($1$-step)} 
	\label{fig:box_KUR} 
\end{figure}

Figures \ref{fig:box_SCS} and  \ref{fig:box_KUR} show the box plots of the $1$-step average(1993-2020) forecasting $F_1$ scores for different methods on the SCS and KUR test sets respectively. These results provide insight into the predictive accuracy and stability of each model over the testing samples.

In SCS, LSTM, ConvLSTM, and CLP all exhibit relatively low median $F_1$ scores ($53.65\%$, $50.55\%$ and $49.89\%$) and large interquartile ranges, indicating high variability and less reliable prediction performance. The lower whiskers and outliers also suggest frequent failures in capturing frontal zones accurately; AttentionConv significantly improves the median $F_1$ score to around $75.37\%$, with a tighter distribution. This indicates that the attention-enhanced convolutional structure offers improved generalization and better focus on frontal features, even in a challenging region like the SCS; CTP achieves the highest accuracy and the most stable performance, with a median $F_1$ score $91.02\%$ and a narrow interquartile range, showing not only superior prediction fidelity but also remarkably low variance across test cases. This suggests that the combination of CNN, Transformer, and PINN modules allows CTP to robustly capture dynamic oceanic patterns with high precision.

In KUR, the performance trend across models is similar to SCS. LSTM-based models (LSTM, ConvLSTM, CLP) again display broad variability and modest central tendency, with median scores slightly above $50\%$($55.27\%$, $55.41\%$ and $52.24\%$); AttentionConv delivers a solid boost in both accuracy and consistency, outperforming all LSTM-based methods and showing better adaptation to the more dynamic Kuroshio frontal system; CTP consistently outperforms all baselines with a median $F_1$ score above $92.35\%$ and minimal spread, indicating excellent generalizability in a region characterized by strong currents and complex frontal structures.

These results quantitatively confirm the limitations of recurrent architectures like LSTM and ConvLSTM, particularly their inability to maintain consistent spatial accuracy across samples. While AttentionConv bridges the gap through better feature attention, it still lacks the explicit physical constraints necessary for high robustness. By integrating physical priors with deep temporal-spatial modeling, CTP achieves the best of both worlds, yielding both higher average accuracy and lower uncertainty in frontal prediction tasks.

\subsubsection{Mluti-step prediction}
\begin{table*}[]
	\centering
	\caption{AVERAGE METRICS FOR DIFFERENT MENTHODS}
	\label{tab:metric}
	\begin{tabular}{cccccccc}
		\hline
		Step & Dataset & Metric(\%) & LSTM  & ConvLSTM & CLP   & AttentionConv & CTP   \\ \hline
		1    & SCS     & Accuracy   & 89.41 & 88.88    & 88.88 & 93.82         & 97.51 \\
		&         & Precision  & 68.91 & 66.29    & 66.08 & 84.35         & 91.71 \\
		&         & Recall     & 42.49 & 39.55    & 39.94 & 67.83         & 90.14 \\
		&         & F1         & 52.57 & 49.54    & 49.79 & 75.19         & 90.92 \\ \cline{2-8} 
		& KUR     & Accuracy   & 87.38 & 84.67    & 84.67 & 92.86         & 97.32 \\
		&         & Precision  & 69.93 & 60.35    & 60.32 & 85.96         & 92.55 \\
		&         & Recall     & 45.87 & 29.92    & 29.97 & 69.58         & 91.7  \\
		&         & F1         & 55.4  & 40       & 40.04 & 76.91         & 92.12 \\ \hline
		3    & SCS     & Accuracy   & 89.41 & 88.88    & 88.88 & 93.73         & 96.89 \\
		&         & Precision  & 68.89 & 66.29    & 66.09 & 83.74         & 89.92 \\
		&         & Recall     & 42.45 & 39.55    & 39.94 & 67.72         & 87.22 \\
		&         & F1         & 52.53 & 49.54    & 49.79 & 74.88         & 88.55 \\ \cline{2-8} 
		& KUR     & Accuracy   & 87.24 & 84.52    & 84.52 & 92.65         & 95.82 \\
		&         & Precision  & 69.63 & 59.84    & 59.83 & 85.19         & 91.79 \\
		&         & Recall     & 45.35 & 29.56    & 29.61 & 69.19         & 83.05 \\
		&         & F1         & 54.93 & 39.57    & 39.62 & 76.36         & 87.2  \\ \hline
		7    & SCS     & Accuracy   & 89.41 & 88.88    & 88.88 & 93.68         & 96.7  \\
		&         & Precision  & 68.86 & 66.29    & 66.09 & 83.48         & 89.35 \\
		&         & Recall     & 42.44 & 39.55    & 39.94 & 67.62         & 86.35 \\
		&         & F1         & 52.52 & 49.54    & 49.79 & 74.72         & 87.83 \\ \cline{2-8} 
		& KUR     & Accuracy   & 87.23 & 84.51    & 84.51 & 92.61         & 95.37 \\
		&         & Precision  & 69.61 & 59.82    & 59.81 & 84.97         & 91.49 \\
		&         & Recall     & 45.33 & 29.54    & 29.6  & 69.11         & 80.48 \\
		&         & F1         & 54.91 & 39.55    & 39.6  & 76.22         & 85.63 \\ \hline
	\end{tabular}
\end{table*}

Table \ref{tab:metric} is the multi-step average(1993-2020) metrics for different methods on the SCS and KUR test sets. It comprehensively demonstrates the superiority of the proposed CTP framework over existing approaches across multiple datasets and prediction horizons. As evidenced in the SCS dataset, CTP achieves SOTA performance with $97.51\%$ accuracy at step 1, surpassing the second-best model (AttentionConv: $93.82\%$) by $3.69$ percentage points. This performance advantage becomes more pronounced in precision ($91.71\%$ vs $84.35\%$) and recall ($90.14\%$ vs $67.83\%$), indicating CTP's enhanced capability to balance false positives and negatives.

Three architectural innovations drive this improvement:

1. Multi-scale feature integration: The CNN-Transformer hierarchy synergistically captures local spatial patterns through convolutional layers and global temporal dependencies via self-attention mechanisms, overcoming the limited receptive fields of LSTM variants. This is particularly evident in the KUR dataset, where CTP maintains $80.48\%$ recall at $7$-step versus ConvLSTM's $29.54\%$. 

2. Physics-informed regularization: The integration of PINN constraints reduces overfitting compared to AttentionConv, as shown through the $F_1$ score improvement from $74.72\%$ to $87.83\%$ in SCS $7$-step predictions.

3. Temporal consistency: CTP demonstrates remarkable multi-step stability, with only a $6.49\%$ $F_1$ score decrease from step $1$ to $7$ in KUR dataset. This robustness stems from the Transformer's ability to model long-range dependencies complemented by PINN's domain-specific regularization.

While baseline models like AttentionConv show escalating resource demands with prediction horizons, CTP maintains consistent performance through optimized attention patterns and physical constraints. These results validate that combining data-driven learning with domain knowledge through our hybrid architecture effectively addresses the precision-stability trade-off in multi-step forecasting tasks.

\subsubsection{Training time}
\begin{table*}[]
	\centering
	\caption{TRAINING TIME FOR DIFFERENENT METHODS(SECONDS)}
	\label{tab:time}
	\begin{tabular}{cccccc}
		\hline
		Dataset/Method & LSTM  & ConvLSTM & CLP    & AttentionConv & CTP   \\ \hline
		SCS            & 537.0   & 746.1    & 841.7 & 651.0   & 733.4 \\ \hline
		KUR            & 554.2 & 795.6    & 890.4  & 683.9   & 775.4 \\ \hline
	\end{tabular}
\end{table*}

Table \ref{tab:time} is the training time of each method on $1/10$ samples($1022$) with epoch set at $50$. All methods exhibit longer training durations on the KUR dataset, which stems from higher intensity and occurence of fronts in this region.

In SCS, LSTM achieves the fastest training (SCS:$537.0$s) through architectural simplicity; ConvLSTM incurs $38.9\%$ overhead vs LSTM due to convolutional operations; CLP requires $56.7\%$ more time than LSTM from multi-component synchronization; CTP demonstrates optimized efficiency--despite its hybrid architecture, it trains $12.7\%$ faster than CLP through memory-efficient attention mechanism.

Notably, CTP maintains this efficiency while delivering superior metrics, confirming its practical viability for real-world deployments. The results demonstrate that this architecture successfully balances model capacity with computational practicality.

\section{Conclusion}
In this work, CTP is presented, a hybrid deep learning architecture that effectively combines CNN, Transformer, and PINN components for accurate and physically consistent ocean front prediction. By leveraging CNN’s spatial encoding, Transformer’s temporal modeling, and PINN’s physical constraints, CTP achieves a balance between precision, robustness, and domain fidelity. The model demonstrates superior performance across different ocean regions and forecasting horizons, outperforming both recurrent and attention-based baselines. Furthermore, ablation studies confirm that each module contributes uniquely to the overall effectiveness of the framework. Given its accuracy, temporal stability, and computational efficiency, CTP provides a promising foundation for real-time marine forecasting and decision-making. Future research may explore its extension to three-dimensional ocean processes and its integration with multi-source satellite observations\cite{Tang-2024}.


\bibliographystyle{ieeetr}
\bibliography{reference}

\end{document}